\documentclass{article}
\usepackage{spconf}
\usepackage{graphics}
\usepackage{amsmath}
\usepackage{amsfonts}
\usepackage{amssymb}

\usepackage{amsbsy}        
\usepackage{nicefrac}      

\usepackage{hyperref}

\usepackage{graphicx}

\newcommand{\bdelta}{\pmb{\delta}}     
\newcommand{\refeq}[1]{Eq.{$\;$}(#1)}  
\newcommand{\Fg}[1]{Fig. #1}           
\newcommand{\coma }{,}
\newcommand{\punto}{.}

\title{Assisted Dictionary Learning for fMRI data analysis}

\name{Manuel Morante Moreno${}^ {1,2}$, Yannis Kopsinis${}^{2,3}$, Eleftherios Kofidis${}^{4,2}$, Christos Chatzichristos${}^{1,2}$,\vspace{-3mm}}
\address{\vspace{-3mm} \emph{Sergios Theodoridis}${}^{1,2,5}$ \vspace{5mm} \\${}^ {1}$ Dept. of Informatics and Telecommunications, University of Athens (Greece), {\small \texttt{morante@cti.gr}} \\${}^{2}$ Computer Technology Institute \& Press ``Diophantus'' (CTI), Patras (Greece), {\small \texttt{chatzichris@cti.gr}} \\ ${}^{3}$ LIBRA MLI Ltd, Edinburgh (UK), {\small \texttt{kopsinis@ieee.org}} \\ ${}^{4}$ Dept. of Statistic and Insurance Science, University of Piraeus, Piraeus (Greece), {\small \texttt{kofidis@unipi.gr}} \\ ${}^{5}$ IAASARS, National Observatory of Athens, GR-15236, Penteli (Greece), {\small \texttt{stheodor@di.uoa.gr}}\vspace{-5mm}
\thanks{\scriptsize The research leading to these results has received funding from the European Union's H2020 Framework Programme (H2020-MSCA-ITN-2014) under grant agreement n° 642685 MacSeNet.}}

\begin{document} 

\maketitle

\begin{abstract}
Extracting information from functional magnetic resonance (fMRI) images has been a major area of research for more than two decades. The goal of this work is to present a new method for the analysis of fMRI data sets, that is capable to incorporate a priori available information, via an efficient optimization framework. Tests on synthetic data sets demonstrate significant performance gains over existing methods of this kind.
\end{abstract}


\section{INTRODUCTION}
\label{sec:intro}
Functional magnetic resonance imaging (fMRI) is a powerful non-invasive technique suitable to providing important information concerning the brain activity. Studying the different areas in the brain that correspond to important tasks such as vision, perception, recognition, etc., constitutes a major open area of research, demanding robust and high precision techniques for the analysis of fMRI data analysis \cite{M-Lee2016-Sparse}, \cite{P-Pro2016-Evaluating}, \cite{B-The2015-MLB}, \cite{G-Bri2015-If`}.

Such data are generated as, a sequence of 3D images of the brain, which are successively acquired along time. Each one of these images is formed by the concatenation of elementary cubes, called \emph{voxels}. Accordingly, the value of each voxel reflects the degree of activity in a certain brain spot. Each 3D image is unfolded into a large row vector, $\mathbf{x}=\left[x_{1},x_{2},\ldots,x_{N}\right]\in\mathbb{R}^{N}$, where $N$ is the total number of voxels. Then, all such data vectors are concatenated together to form the data matrix, $\mathbf{X}\in\mathbb{R}^{T\times N}$, where $T$ is the total number of successive time acquisitions.

In the brain, a number of different functions/processes take place simultaneously; thus, the obtained data consists of a mixture of various activation signals referred to as \emph{sources}. The aim of fMRI data analysis is to unmix those sources in order to reveal both their activation patterns as well as the corresponding activated brain areas, associated with each one of the sources.

From a mathematical point of view, the source unmixing task can be described as a problem of factorization of the data matrix, i.e.,
\begin{equation} \label{eq:DL}
	\mathbf{X}\thickapprox\mathbf{DS}\coma
\end{equation}
where $\mathbf{D}\in\mathbb{R}^{T\times K}$ is a matrix, whose columns represent the activation patterns or time courses, associated with each one of the sources, $\mathbf{S}\in\mathbb{R}^{K\times N}$ is the matrix whose rows model the brain areas, activated by the corresponding sources, and $K$ is the number of sources, whose value is set by the user. The rows of the matrix $\mathbf{S}$ are usually referred to as spatial maps.

fMRI essentially measures the changes in the level of oxygen in blood caused by the neural activity, which leads to an indirect measure of the latter. More specifically, the observed/measured signal results from the convolution of the true activations with the, so called, Hemodynamic Response Function (HRF). HRF varies across different persons as well as across different brain areas of the same person \cite{H-Agu1998-vhB}.

A widely used  tool in fMRI analysis is the General Linear Model (GLM), which relies on the assumed form of the HRF in order to construct the matrix $\mathbf{D}$ in \refeq{\ref{eq:DL}}. In particular, the specific design of each experiment allows to make a guess of the true time instances, where the activations are expected to appear. Adopting a functional form for HRF and convolving it with the expected activation sequence, the time course corresponding to the specific task can be estimated and considered known. Hereafter, such estimated time courses are referred to as \emph{task-related} time courses.

Alternatively, one might use a blind source separation (BSS) approach, which can simultaneously estimate $\mathbf{D}$ and $\mathbf{S}$ without having a resort to any assumptions regarding HRF. To this end, different assumptions concerning either statistical or structural properties of the involved matrices are adopted. Namely, Independent Component Analysis (ICA) \cite{M-McK1998-Afd}, \cite{f-Cal2006-Ufw}, \cite{G-Gri2014-Iba}, which has been widely used in the fMRI unmixing problem, assumes independence among the sources, whereas Dictionary Learning (DL)-based techniques \cite{M-Kop2014-FMRI}, which have been gaining more attention recently, exploit the fact that the matrix $\mathbf{S}$ is expected to be sparse. This is true, since the brain can be considered as a \emph{sparse system}; each task/function produces an activation pattern which appears localized in specific regions \cite{B-Val2005-Ebfcwsma}.

Recently, a method , called \emph{Supervised Dictionary Learning} (SDL) \cite{M-Zha2015-SDL}, which allows the incorporation of information related to the HRF in a BSS framework, was presented, leading to enhanced results. However, both GLM and SDL suffer from the same shortcoming; that is, an accurate enough assumption about the functional form of the HRF needs to be made.

In this paper, a new DL method is proposed, which, although it exploits the benefits of incorporating some a priori knowledge concerning the HRF, allows for substantial tolerance against inaccurate choices of its respective form.

\section{ Assisted DL for \lowercase{f}MRI Data Analysis}

\subsection{Supervised Dictionary Learning}
In practice, the use of estimated time courses generally provides satisfactory results in GLM. On the other hand, the assumption concerning the spatial sparsity encapsulates highly valuable information, which is mathematically translated to the equivalent use of sparsity constraints in DL methods. However, it turns out that the a priori assumptions, adopted by these methods, are not by themselves powerful enough to come up with sufficiently good results.

The starting point in the formulation of the SDL, lies in the splitting of the main dictionary in two parts:
\begin{equation}
	\mathbf{D}=\left[\mathbf{\Delta},\mathbf{D}_{F}\right]\in\mathbb{R}^{T\times K}\coma
	\label{eq:KorDivision}
\end{equation}
where the first part, $\mathbf{\Delta}\in\mathbb{R}^{T\times M}$, is constrained to contain the imposed task-related time courses and is considered fixed. In contrast, the second part, $\mathbf{D}_{F}\in\mathbb{R}^{T\times(K-M)}$, is the variable one to be estimated via DL optimizing arguments.

The result is still a DL scheme but it incorporates specific time courses; it turns out that the reported results lead to an enhanced performance, compared to those obtained via the use of a standard DL technique. Nevertheless, this approach still inherits the same major drawback associated with GLM. That is, the constrained dictionary atoms (columns of the $\mathbf{\Delta}$ matrix) only help if the a-priori imposed information is accurate enough; however, if the imposed time courses are shifted or miss-modelled, their contribution can have a detrimental effect, leading to wrong results. 

\subsection{Atom-Assisted DL}  
In this paper, an alternative approach is presented, that provides a more relaxed way of incorporating the a-priori adopted forms of the time courses. The main idea is to consider that the atoms of the constrained part are not necessarily equal to the a-priori selected ones; instead, a looser constraint is employed, embedded in the optimization process. Thus, the strong \emph{equality} demand is relaxed by a looser \emph{similarity} distance-measuring norm constraint. 

Hence, if part of the a priori information is not accurate enough, since the constrained atoms are not considered fixed any more, the method is free to readjust them, with respect to the information that resides in the data, in an optimum way. It turns out that such an approach robustifies the procedure against the major drawback associated with the HRF-based methods.

The starting point is, again, to split the dictionary:
\begin{equation}
	\mathbf{D}=\left[\mathbf{D}_{C},\mathbf{D}_{F}\right]\in\mathbb{R}^{T\times K}\punto
	\label{eq:DecMoM}
\end{equation}
In contrast to the SDL approach, however, the constrained part, $\mathbf{D}_{C}\in\mathbb{R}^{T\times M}$, is not considered fixed any more; instead, it can vary in line with the constrained optimization cost.

The optimization task, adopted here, is formulated as:
\begin{equation} \label{eq:OptTask}
	(\hat{\mathbf{D}},\hat{\mathbf{S}})=\underset{\mathbf{D},\mathbf{S}}{\text{argmin}}\left\Vert \mathbf{X}-\mathbf{D}\mathbf{S}\right\Vert _{F}^{2}+\lambda\left\Vert \mathbf{S}\right\Vert _{1,1}\:\text{s.t.}\:\mathbf{D}\in\mathfrak{D} 
\end{equation}
where, $\left\Vert \mathbf{S}\right\Vert _{1,1}=\sum_{i}^{K}\sum_{j}^{N}|s_{ij}|$ is the sparsity-promoting term over the coefficient matrix, and $\mathfrak{D}$ is an admissible set of dictionaries. In our case, $\mathfrak{D}$ comprises the set of dictionaries sharing the following property:  
\begin{equation}
\mathfrak{D}=\left\{ \mathbf{D}\in\mathbb{R}^{T\times K}\,:\,\begin{array}{l}
\left\Vert \mathbf{d}_{i}-\bdelta_{i}\right\Vert ^{2}_{2}\leqslant c_{\delta},\:\forall i\in[1,M]\subset\mathbb{N}\\
\left\Vert \mathbf{d}_{i}\right\Vert ^{2}_{2}\leqslant c_{d},\:\forall i\in[M+1,K]\subset\mathbb{N}
\end{array}\right\} \coma
\end{equation}
where $\mathbb{N}$ is the set of natural numbers, $\left\Vert \cdot\right\Vert _{2}$ denotes the 2-norm, $\mathbf{d}_{i}$ is the $i^{\mathrm{th}}$ column of the dictionary $\mathbf{D}$ and $\bdelta_{i}$ is the $i^{\mathrm{th}}$ a-priori selected task-related time course. The constant $c_{\delta}$ is a user-defined parameter which controls the \emph{degree of similarity} between the constrained atoms and the imposed time courses. The remaining dictionary atoms are constrained to have a bounded norm no larger than a prefixed parameter $c_d$.

\subsection{Optimization Method}
In order to solve the previous optimization task, the majorization method \cite{O-Yag2009-Dictionary} is adopted, which has already been used in the past to solve DL problems. No doubt, any other relevant optimization method can be mobilized, and its most appropriate choice is currently under study. Although the adoption of the majorization method does not require to implement a Lagrangian relaxation of the constrained problem it is considered for simplicity. Thus, the equivalent optimization task, via the corresponding Lagrangian formulation of the minimization problem in \refeq{\ref{eq:OptTask}}, is given by
\begin{gather}
	(\hat{\mathbf{D}},\hat{\mathbf{S}})=\underset{\mathbf{D},\mathbf{S}}{\text{argmin}}\;\phi_{\lambda,\mathbf{\gamma}}(\mathbf{D},\mathbf{S}),\quad\text{where} \\ \label{eq:optcostA}
	\phi_{\lambda,\mathbf{\gamma}}(\mathbf{D},\mathbf{S})=\left\Vert \mathbf{X}-\mathbf{D}\mathbf{S}\right\Vert _{F}^{2}+\lambda\left\Vert \mathbf{S}\right\Vert _{1,1}+\mathcal{P}_{\mathbf{\gamma}}(\mathbf{D}).
\end{gather}
$\mathcal{P}_{\mathbf{\gamma}}(\mathbf{D})$ depends on the dictionary and is defined as
\begin{equation} \label{eq:WholeTerm}
	\begin{split}
	\mathcal{P}_{\mathbf{\gamma}}(\mathbf{D}) = & \sum_{i=1}^{M}\gamma_{i}\left[\left(\mathbf{d}_{i}-\bdelta_{i}\right)^{T}\left(\mathbf{d}_{i}-\bdelta_{i}\right)-c_{\delta}\right] + \\ 
	& +\sum_{i=M+1}^{K}\gamma_{i}\left(\mathbf{d}_{i}^{T}\mathbf{d}_{i}-c_{d}\right)\coma
	\end{split}
\end{equation}
where the introduced parameters, $\gamma_{i}$, $i=1,2,\ldots,K$ correspond to the associated $K$ Lagrangian multipliers.

Eq. \eqref{eq:WholeTerm} can be compactly expressed as:
\begin{equation}
	\mathcal{P}_{\mathbf{\Gamma}}(\mathbf{D})=\text{tr}\left[\mathbf{\Gamma}\left(\mathbf{D}- \mathbf{\Delta}\mathbf{M}\right)^{T}\left(\mathbf{D}-\mathbf{\Delta} \mathbf{M}\right)-\mathbf{C}\right]\coma
\end{equation}
where $\mathbf{M}\in\mathbb{R}^{M\times K}$, has zero values everywhere but in $M_{ij}$, with $i=j$, which equal to one,
$\mathbf{\Gamma}$ is a diagonal matrix with $\gamma_{i}$,\,$i=1,2,\ldots,K$ as the $i^{th}$ diagonal element and $\mathbf{C}$ is a diagonal matrix with the corresponding parameters $c_{\delta}$ and $c_{d}$, on its diagonal. Accordingly, the cost function \eqref{eq:optcostA} is rewritten as:
\begin{equation} \label{eq:FinalCostFunction}
		\phi_{\lambda,\mathbf{\Gamma}}(\mathbf{D},\mathbf{S})=\left\Vert \mathbf{X}-\mathbf{D}\mathbf{S}\right\Vert _{F}^{2}+\lambda\left\Vert \mathbf{S}\right\Vert _{1,1}+\mathcal{P}_{\mathbf{\Gamma}}(\mathbf{D})\punto
\end{equation}

\subsection{The Algorithm}
The optimization with respect to $\mathbf{D}$ and $\mathbf{S}$ is a challenging one and is largely simplified by adopting a two-step alternating minimization iterative procedure. In particular, starting from arbitrary estimates, $\mathbf{D}_{(0)}$ and $\mathbf{S}_{(0)}$, the algorithm comprises the following steps:
\begin{align} \label{eq:StepI}
	\textbf{Step I}  && {\displaystyle \min_{\mathbf{S}} \phi_{\lambda,\mathbf{\Gamma}}(\mathbf{D},\mathbf{S})\quad} & \text{fixed }\mathbf{D}\coma \\
	\label{eq:StepII}
	\textbf{Step II} && {\displaystyle \min_{\mathbf{D}} \phi_{\lambda,\mathbf{\Gamma}}( \mathbf{D},\mathbf{S})\quad} & \text{fixed }\mathbf{S}\punto
\end{align}

Following the majorization technique, for each step, the objective function is replaced by a surrogate one, which majorizes it, and is easier to be iteratively minimized, compared to the original one. The surrogate function is not unique, but it has to satisfy specific conditions, e.g., \cite{O-Yag2009-Dictionary}.

\subsubsection{Step I: Coefficient Update}
At the $t^{\mathrm{th}}$ step of the alternating minimization of \refeq{\ref{eq:StepI}}, the objective function is minimized with respect to $\mathbf{S}$ keeping $\mathbf{D}$ fixed at its currently available estimate, $\mathbf{D}=\mathbf{D}_{(t)}$. This minimization is also achieved in an iterative way and through the introduction of a surrogate function. Starting the iterations from the currently available estimate, $\mathbf{S}^{[0]}=\mathbf{S}_{(t)}$, the estimate, $\mathbf{S}^{[n]}$, at the $n^{\mathrm{th}}$ iteration, is obtained in terms of the previously estimate of $\mathbf{S}^{[n-1]}$ by minimizing the following surrogate function \cite{O-Yag2009-Dictionary},
\begin{equation} \label{eq:SurrI}
	\psi_{\lambda}(\mathbf{S},\mathbf{S}^{[n-1]})=\phi_{\lambda,\mathbf{\Gamma}}(\mathbf{D},\mathbf{S})+\pi_{S}(\mathbf{S},\mathbf{S}^{[n-1]})\coma
\end{equation}
where \linebreak $\pi_{S}(\mathbf{S},\mathbf{S}^{[n-1]}):=c_{S}\left\Vert \mathbf{S}-\mathbf{S}^{[n-1]}\right\Vert _{F}^{2}-\left\Vert \mathbf{D}\mathbf{S}-\mathbf{D}\mathbf{S}^{[n-1]}\right\Vert _{F}^{2}$ and $c_{S}>\left\Vert \mathbf{D}^{T}\mathbf{D}\right\Vert _{2}$, is a constant where $\left\Vert \cdot\right\Vert _{2}$ is defined as the spectral norm. Thus, two different iterations run in a nested form; for each iteration with respect to $(t)$, there is an (inner) iteration with respect to $[n]$.

Let $\mathbf{A}:=\frac{1}{c_{S}}\left(\mathbf{D}^{T}\mathbf{X}+\left(c_{S}\mathbf{I}_{K}-\mathbf{D}^{T}\mathbf{D}\right) \mathbf{S}^{[n-1]}\right)$. It can be shown that the optimum value of the surrogate function above is found by shrinking the elements in $\mathbf{A}$, that is, 
\begin{gather}
	\mathbf{S}^{[n]}=\mathcal{S}_{\lambda}(\mathbf{A})\coma \quad \text{where} \\ 
	\mathcal{S}_{\lambda}(\mathbf{A})\;:\; s_{ij}=\begin{cases}
a_{ij}-\frac{\lambda}{2}\text{sign}\left(a_{ij}\right) & \text{if }\left|a_{ij}\right|>\frac{\lambda}{2}\\
0 & \text{otherwise}
\end{cases}\punto
\end{gather}
The iterative update continues until a stopping criterion is met. The pseudocode for this coefficient update is presented in \emph{Algorithm 1}.

\subsubsection{Step II: Dictionary Update}
In the second step of the alternating minimization, the objective function is minimized with respect to $\mathbf{D}$, keeping $\mathbf{S}$ fixed at its currently available estimate, $\mathbf{S}=\mathbf{S}_{(t+1)}$. A majorization rationale is also used for this step as well.

To this end, an appropriate surrogate function is introduced given by
\begin{equation} \label{eq:SurrII}
	\psi_{\mathbf{\Gamma}}(\mathbf{D},\mathbf{R})=\phi_{\lambda,\mathbf{\Gamma}}(\mathbf{D},\mathbf{S})+\pi_{D}(\mathbf{D},\mathbf{R})\coma
\end{equation}
where $c_{D}>\left\Vert \mathbf{S}^{T}\mathbf{S}\right\Vert _{2}$ is a constant and $\mathbf{R}=\mathbf{D}^{[n-1]}$ is the estimate of the dictionary of the previous step.

Minimizing \refeq{\ref{eq:SurrII}} with respect to $\mathbf{D}$ takes place also iteratively, starting from $\mathbf{D}^{[0]}=\mathbf{D}_{(t)}$. The optimum value of the surrogate function is found at the point of zero gradient:
\begin{equation}
	\frac{d\psi_{\mathbf{\Gamma}}}{d\mathbf{D}}=-2\mathbf{X}\mathbf{S}^{T}+2\left(\mathbf{D}- \mathbf{\Delta}\mathbf{M}\right)\mathbf{\Gamma} +2c_{D}\left(\mathbf{D}-\mathbf{R}\right) +2\mathbf{R}\mathbf{S}\mathbf{S}^{T}\punto
\end{equation}
\begin{table}
\begin{tabular}{cl}
 &  \emph{Algorithm 1 - Step I}\tabularnewline
\hline 
\vspace{-3mm} & \tabularnewline 
1: & \textbf{Initialization:} $c_{S}>\left\Vert \mathbf{D}^{T}\mathbf{D}\right\Vert _{2}$,
$\mathbf{S}^{[0]}=\mathbf{S}_{(t)}$, $n=0$\tabularnewline
2: &  \textbf{repeat}  \tabularnewline
3: & $\quad n=n+1$ \tabularnewline
4: & $\quad\mathbf{A}=\frac{1}{c_{S}}\left(\mathbf{D}^{T}\mathbf{X}+\left(c_{S}\mathbf{I}_{K}-\mathbf{D}^{T}\mathbf{D}\right)\mathbf{S}^{[n-1]}\right)$\tabularnewline
5: & $\quad\mathbf{S}^{[n]}=\mathcal{S}_{\lambda}(\mathbf{A})$\tabularnewline
6: & \textbf{until} stop criterion is met${}^{*}$\tabularnewline
7: & \textbf{output:} $\mathbf{S}_{(t+1)}=\mathbf{S}^{[n]}$\tabularnewline
\hline 
\vspace{-2mm} & \tabularnewline
  & \emph{Algorithm 2 - Step II} \tabularnewline
\hline 
\vspace{-3mm} & \tabularnewline
1:  & \textbf{Initialization:} $c_{D}>\left\Vert \mathbf{S}^{T}\mathbf{S}\right\Vert _{2}$,
$\mathbf{D}^{[0]}=\mathbf{D}_{(t)}$, $n=0$\tabularnewline
2: & \textbf{repeat}\tabularnewline
3: & $\quad n=n+1$ \tabularnewline
4:  & $\quad\mathbf{B}=\frac{1}{c_{D}}\left(\mathbf{X}\mathbf{S}^{T}+\mathbf{R}\left(c_{D}\mathbf{I}_{K}-\mathbf{S}\mathbf{S}^{T}\right)\right)$\tabularnewline
5: & $\quad\mathbf{D}^{[n]}=\mathcal{U}(\mathbf{B})$\tabularnewline
6: & \textbf{until} stop criterion is met${}^{*}$\tabularnewline
7: & \textbf{output:} $\mathbf{D}_{(t+1)}=\mathbf{D}^{[n]}$\tabularnewline
\hline 
\end{tabular}

$\quad {}^{*}$ {\footnotesize In this paper, a fixed number of iterations is used.}
\end{table}

Setting the derivative above equal to zero, solving for $\mathbf{D}$ and setting $\gamma_i$ to values that satisfy the Karush-Kuhn-Tucker (KKT) conditions (details are omitted due to lack of space), a two-step procedure for the dictionary update results, following arguments similar to those in \cite{O-Yag2009-Dictionary}. Namely, an intermediate  quantity $\mathbf{B}$ is first defined and computed as $\mathbf{B}:=\frac{1}{c_{D}}\left(\mathbf{X}\mathbf{S}^{T}+\mathbf{R}\left(c_{D}\mathbf{I}- \mathbf{S}\mathbf{S}^{T}\right)\right)$.

The estimates of the updated dictionary atoms are then given by
\begin{equation}
\mathbf{d}_{i}^{[n]}=\begin{cases}
{\scriptstyle \text{for }i\in\left[1,M\right]}\,,\quad\begin{cases}
\mathbf{b}_{i} & \text{if }\left\Vert \mathbf{b}_{i}-\bdelta_{i}\right\Vert ^{2}_{2}\leqslant c_{\delta}\\
\frac{c_{\delta}\left(\mathbf{b}_{i}-\bdelta_{i}\right)}{\left\Vert \mathbf{b}_{i}-\bdelta_{i}\right\Vert ^{2}}+\bdelta_{i} & \text{otherwise}
\end{cases}\\
\\
{\scriptstyle \text{for }i\in[M+1,K]}\,,\,\begin{cases}
\mathbf{b}_{i} & \text{if }\left\Vert \mathbf{b}_{i}\right\Vert ^{2}\leqslant c_{d}\\
\frac{c_{d}}{\left\Vert \mathbf{b}_{i}\right\Vert ^{2}}\mathbf{b}_{i} & \text{otherwise}
\end{cases}
\end{cases}\coma
\end{equation}

It is not difficult to show that after this update, the KKT conditions for all dictionary columns hold. Moreover, the set of all admissible dictionaries, $\mathfrak{D}$, constitutes a convex non-empty set. It can be shown that this fact guarantees that the proposed algorithm converges for random initialization. Due to the space limitations imposed by a conference paper, details are omitted. The pseudo-code for this dictionary update is presented in \emph{Algorithm 2}. Furthermore, the MATLAB code for this method can be freely downloaded from {\small \url{https://github.com/MorCTI/Atom-Assisted-DL.git}}.

\begin{figure}[b] 
	\vspace{-3mm}
	\centering
	\includegraphics[width=90mm]{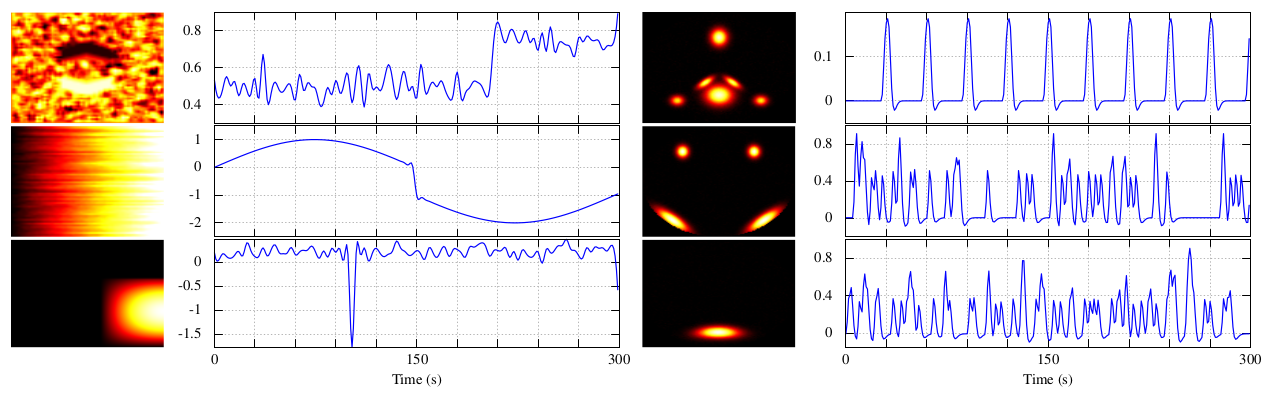}
	\vspace{-8mm}
	\caption{Selection of different simulated sources. In the first column, Gaussian, subgaussian and supergaussian artefacts are plotted from the artificial data set in \cite{O-Cor2005-Comparison}. In the second column, three other simulated yet realistic brain sources are shown. The first one corresponds to the source of interest.}
    \label{fig:Patterns}
\end{figure}

\begin{figure}[th]
	\vspace{-5mm}
	\centering
	\includegraphics[width=1.\columnwidth]{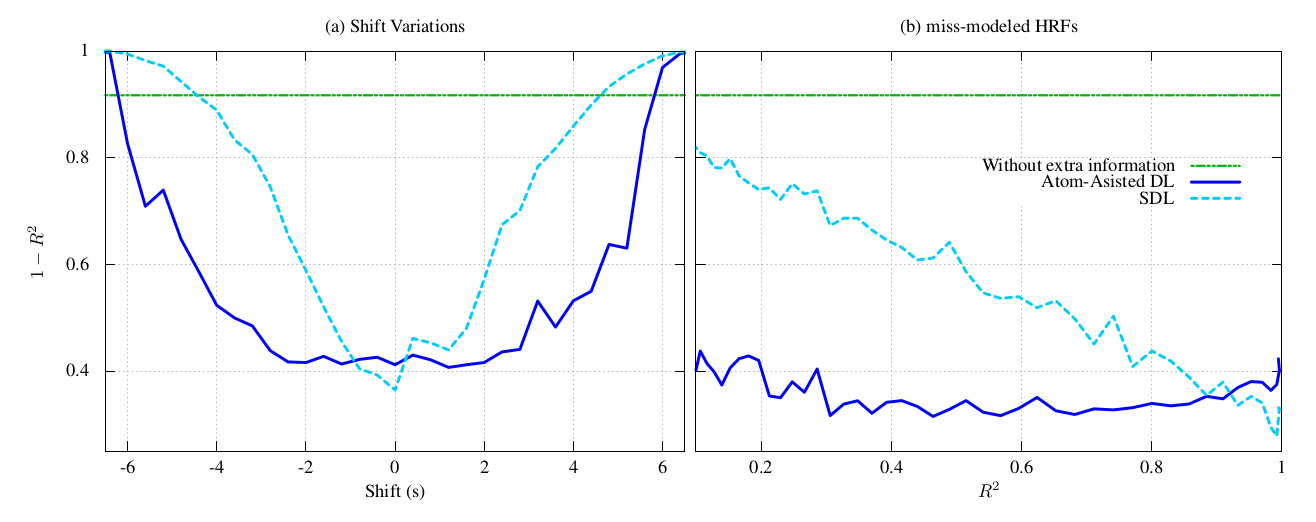}
	\vspace{-3mm}
	\caption{Squared correlation coefficient between the estimated source and the true one for the two miss-modelling experiments.}
	\label{fig:CurvErr}
\end{figure}

\vspace{-2mm}
\section{Performance Evaluation} 
\vspace{-2mm}
The aim of this section is twofold. First, to demonstrate the advantages obtained by incorporating the external information about task-related time courses. Second, to compare the sensitivity of the proposed scheme with that of SDL, in cases where the imposed time courses  deviate from the true ones.

The data set used is synthetic and generated with SimTB\footnote{SimTB simulator is a free MATLAB toolbox available for download in ({\small \url{http://mialab.mrn.org/software}}), which has been lately adopted in a number of fMRI data analysis studies, e.g.  \cite{M-All2012-Cis},\cite{M-Ma2011-Aif},\cite{M-Ma2013-Cgv},\cite{M-Ge2015-tss}.} \cite{M-Erh2012-Sst}. In order to make the data more realistic, the sources 3, 4, 5, 7, 8 of the data set in \cite{O-Cor2005-Comparison}, which represents machine artifacts, are also added. The data set used can be downloaded from ({\small \url{https://github.com/MorCTI/Atom-Assisted-DL.git}}). As an example, in \Fg{\ref{fig:Patterns}}, as an example, 6 among the 20 sources used in total are depicted. Note that the current performance study cannot be realized based on real fMRI data, since in such a case the ground truth is not known.

With respect to the SDL and atom-assisted DL methods, the larger the number of time-courses which are imposed as constraints in the algorithm, the best is the performance observed, due to the fact that a larger amount of information is provided. Therefore, in order to make things harder, in the evaluation tests that follow, only task-related time course is considered. Moreover, two different miss-modeling cases are examined. In the first one, the task-related time course, is a time-shifted version of the true one. The result is shown in \Fg{\ref{fig:CurvErr}a}, where the solid and the dashed lines correspond to the atom-assisted DL and the SDL, respectively. The horizontal axis represents the time shifting of the imposed task-related time course in relation to the true one, expressed in seconds. The vertical axis shows $1-R^{2}$, with  $R$ being the correlation coefficient between the estimated and the true source.
 It is apparent that the proposed scheme offers enhanced robustness allowing time discrepancies up to 4 seconds (2 seconds in each directions) without any performance degradation. If some performance loss is allowed, 8 seconds of time shift are well tolerated. 
 
The flat dot-dashed curve corresponds to the fully blind approach, i.e., when no information regarding the task-related time course is provided. In this case, the fully blind approach fails to estimate the signal of interest. This happens since in the experimental setup it has been provisioned that the signal of interest a) exhibit significant space overlap with artefact sources and other physical sources and b) have overall energy not higher than of its neighbouring sources. This design generates a hard but realistic experimental scenario, in which other conventional blind source separation methods, such as ICA \cite{f-Cal2006-Ufw} or k-SVD \cite{M-Kop2014-FMRI}, fail to recover the source of interest. The latter was confirmed with various simulation studies which will be presented elsewhere due to space limitations.

Note that both in the current and in the next experiment, all curves result from the ensemble average of 20 independent runs.
Besides, the majorization optimization approach was also used in the SDL case for the dictionary learning task substituting the online DL optimization, \cite{SPAMS}, used in the original paper. This was done in order to guarantee a fair comparison with the proposed approach. In any case, it was verified with extensive simulations that the two optimization approaches resulted in similar performance. 
Moreover, in the atom-assisted DL case, the parameters, $c_{\delta}$, $c_{d}$, $\lambda$ were set equal to 0.2, 1, 0.1, respectively. Moreover, 100 iterations were performed for Algorithms 1 and 2. Finally, 500 alternating minimization iterations were used in all cases.

In the second miss-modeling scenario, shown in \Fg{\ref{fig:CurvErr}b}, the imposed time course results from the convolution of the experimental task event with an HRF which is different from the true one. For the construction of the different tested HRFs, the canonical HRF model is adopted \cite{H-Han2004-VBh}. In order to perform this study, the free parameters of the canonical HRF model are gradually modified leading to HRFs with a successively narrower shape compared to the true HRF. In particular, the horizontal axis shows the squared correlation coefficient, $R^{2}$, between the true HRF and the modified HRF of the corresponding imposed time course. Again, the vertical axis shows $1-R^{2}$, with $R$ being the correlation coefficient between the estimated and the true source. Once again, it is observed that the proposed method is insensitive to large deviations between the provided information and the true one. 

\section{Conclusions} 
In this paper, a new source separation approach for fMRI data analysis is proposed. The method allows for the incorporation of task-related a-priori information which leads to vast performance improvements compared to conventional fully blind approaches. Moreover, the proposed method exhibits enhanced robustness against miss-modelling of the imposed extra information.

\newpage
\bibliographystyle{IEEEbib}
\bibliography{MyBiblio.bbl}

\end{document}